# Parameter Sharing Deep Deterministic Policy Gradient for Cooperative Multi-agent Reinforcement Learning

Xiangxiang Chu, Hangjun Ye
Xiaomi Inc. Beijing, China

## Abstract

Deep reinforcement learning for multi-agent cooperation and competition has been a hot topic recently. This paper focuses on cooperative multi-agent problem based on actor-critic methods under local observations settings. Multi agent deep deterministic policy gradient obtained state of art results for some multi-agent games, whereas, it cannot scale well with growing amount of agents. In order to boost scalability, we propose a parameter sharing deterministic policy gradient method with three variants based on neural networks, including actor-critic sharing, actor sharing and actor sharing with partially shared critic. Benchmarks from rllab show that the proposed method has advantages in learning speed and memory efficiency, well scales with growing amount of agents, and moreover, it can make full use of reward sharing and exchangeability if possible.

## 1. Introduction

Deep neural network and its applications has become a hot research in recent years. It has been applied in a varieties of fields, such as game playing (Mnih et al., 2015; Silver et al., 2016), robot control (Levine et al., 2016), image classification (Krizhevsky et al., 2012), speech recognition (Hinton et al., 2012), machine translation (Girshick et al., 2014), etc.. By combining Q-learning with deep neural network as the value function, a deep Q-network called DQN was put forward to play human-level games just depending on the pixel input, which boosted reinforcement learning to a potential solver for generic artificial intelligence (Mnih et al., 2015). By applying replay memory and target network frozen mechanisms, DQN can behave and converge stably. Afterwards various DQN variants for example, double DQN (Van Hasselt et al., 2016), Duel DQN (Wang et al., 2016), DQN with priority replay memory (Schaul et al., 2015), were proposed and boosted the performance of origin DQN further. All the methods mentioned above belong to value iteration branch in reinforcement learning.

Correspondence to: Xiangxiang Chu <chuxiangxiang@xiaomi.com>.

Another important branch for reinforcement learning is policy gradient (Sutton et al., 2000), which also has a handful of variants. Asynchronous advantage actor critic (A3C) is a fast asynchronously parallel learning method in actor critic family (Mnih et al., 2016), where the actor part outputs policy probability. Unlike A3C's stochastic action policy, deterministic policy gradient using deep neural network (DDPG) as the underling policy model was put forward to solve the continuous control problem (Lillicrap et al., 2015).

All methods above meet with a common thorny problem, policy is usually prone to degrade or even diverge during training. In order to boost the policy monotonically, trust region policy optimization (TRPO) introduced Kullback-Leibler (KL) constraint to control policy change range during one iteration (Schulman, Levine, et al., 2015). Though Hessian free method combining Fish vector multiply and conjugate gradient is used, the computing expense per update is usually one order of magnitude higher than SGD and its variants. Recently, proximal policy optimization (PPO) was proposed to cut down TRPO's computing expense while keeping some benefits from TRPO (Schulman et al., 2017). All these methods mainly concentrate on single agent learning.

Our work focuses on multi-agent cooperation problem. We organize this paper as following. Firstly, we introduce some related work about multi-agent reinforcement learning and essential background knowledge. Secondly, we propose several parameter sharing multi-agent deep deterministic policy gradient variants. Lastly, we test their performance based on three multi-agent benchmarks and end the paper with our future work plan.

## 2. Related Work

Multi-agent learning is still an open research area so far. A traditional approach for multi-agent learning is called IQL and its core idea is taking other agents as the part of environment. It will lead to non-static environment due to ignoring other agents' policies (Tan, 1993). To eliminate the non-static introduced by other agents while keeping and stabilizing the replay memory mechanism from DQN, two methods: important sampling and fingerprints were put forward (Foerster et al., 2017), where the StarCraft micro-battle was used as the test case. However, these two methods have the same limitation as DQN and could only deal with discrete problems.

Recently many contributions focused on extending single agent reinforcement learning to multi-agent based on deep neural networks.

Deep Q-Networks (DQN) was extended into multi-agent DQN to play Pong (Tampuu et al., 2015), where cooperative and competitive scenarios were designed by changing rewards for each agent. Their research focused on discrete action space and global observation for each agent.

TRPO was extended to multi-agent scene using parameter sharing (Gupta et al., 2017), which has some similarities with us. They also focused on cooperative multi-agent RL learning but didn't use actor-critic method.

Deep Deterministic Policy Gradient (DDPG) was extended to multi-agent (Lowe et al., 2017) (MADDPG), which showed that the non-static environment made the multi-agent problems much harder than single agent. Therefore, they utilized a decentralized actor policy network and a centralized critic network for each agent. Input observation for critic network contained local observations and actions from all agents and eliminated the environment non-static. Each agent has a different actor network to act and a different critic network to evaluate. Moreover, they take the decentralized reward in place of traditional centralized global reward. It easily beat various RL methods including DDPG, TRPO, DQN, Actor-Critic and REINFORCE with a large margin on several multi-agent benchmarks. One major and fatal bottleneck of MADDPG is the total number of parameter network grows quadratically as the agent number increases, which is one focus of our work.

## 3. Background

Since our work focuses on improvement for literature (Lowe et al., 2017), we follow their variable naming routine. In fact, a very good brief introduction about the background has been made there, so we choose to omit the repeated background introduction and focus on the different and supplement parts.

### 3.1 Multi-agent Markov Decision Games

Multi-agent Markov games can be defined by $N$ agents with a set of global or local observations $o_1,...,o_N$, a set of actions $A_1,...,A_N$, a set of states $S$ and a state transition function $T: S \times A_1 \times A_2 \times ... \times A_N \mapsto S$ which determines the Markov process. For each agent $i$, it interacts with environment by taking actions following its policy $\pi_{\theta_i}: o_i \times A_i \mapsto [0,1]$, transformed into the next state and get a reward $r_i: S \times A_i \mapsto R$ to judge the policy's performance. Each agent tries to maximize the accumulated discount return $R = \sum_{t=0}^{T} \gamma^t r_i^t$, where T is the expect time horizon and $\gamma$ is the discount parameter. In this paper, only local observations are available for all games.

### 3.2 Q-Learning and DQN

Q learning is a traditional value iteration reinforcement learning method, which updates single Q value based on Bellman equation. We can write Q and V function as following:

$$Q(s,a) = \sum_{t=t_0}^{T} r_t \gamma^{t-t_0} \mid s_{t_0} = s, a_{t_0} = a$$
$$V(s) = \sum_{t=t_0}^{T} r_t \gamma^{t-t_0} \mid s_{t_0} = s \quad (1)$$

Tabular methods are often used to solve simple problems with finite states. Recently, value functions estimation were widely used to boost the iteration efficiency and solve complex problems. The iteration formula can be written as:

$$Q(s,a) = r + \gamma Q(s',a')$$
$$a' = \arg\max_a Q(s',a) \quad (2)$$

The use of value function is not limited to the specific function form, and function forms that can imitate arbitrary complexity are recommend to solve various problems, among which deep neural network is a popular choice. Two mechanisms contribute to DQN's stability: replay memory and target network holding. The state correlations are weaken by uniformly sampling from the replay memory. The temporal difference(TD) loss function can be written as:

$$E_{s,a,r,s'}(Q(s,a|\theta) - y)^2$$
$$y = r + \gamma Q(s',a'|\theta')$$
$$a' = \arg\max_a Q(s',a|\theta') \quad (3)$$

Where $\theta$ represents the current $Q$ function parameter, and $\theta'$ stands for the frozen target $Q$ function parameter. $\theta'$ is updated as to $\theta$ with fixed frequency.

### 3.3 Policy Gradient

As a widely used reinforcement algorithm, policy gradient theorem lays a good mathematical foundation for this branch. Policy gradient gets parameter updated directly from increasing expected Q value function. The gradient formula can be written as (Sutton et al., 2000):

$$\nabla_\theta J_\theta = E_{s \sim p^\pi, a \sim \pi}[\nabla_\theta \log \pi_\theta(a|s) * Q^\pi(s,a)]. \quad (4)$$

$\pi_\theta(a|s)$ is the output policy probability distribution for action $a$. In dealing with continuous actions, actual actions are usually sampled from normal distribution.

Policy gradient has an intuitive form that increases the probability of those actions that have high Q values and others' vice versa. A series of algorithms were proposed based on policy gradient theorem, among which REINFORCE is a famous one (Williams, 1992). Q value function introduces large variance since different states' values may vary greatly. A recommended and widely used solution is to use the difference between Q value and a baseline function $B(s)$ instead, which only correlates with state but not with action, and the V value function is a common choice. This variant policy gradient can be written as:

$$\nabla_\theta J_\theta = E_{s \sim p^\pi, a \sim \pi}[\nabla_\theta \log \pi_\theta(a|s) \\ *(Q^\pi(s,a) - B(s))]. \quad (5)$$

### 3.4 Actor-Critic

A common drawback of PG variants is that value function cannot be obtained until current episode ends, thus resulting in inefficient parameter update. In order to improve iteration efficiency, value function is estimated by critic network. The policy network can be viewed as actor network, so this method branch is called Actor-Critic (Sutton et al., 1998).

The actor network gradient is the same as (5), and the critic loss function is usually but not limited to TD value loss. Without loss of generality, the critic loss function based on $V$ value function can be written as:

$$E_{s,r,s'}[V_\theta(s) - (r + \gamma V_{\theta'}(s'))]^2. \quad (6)$$

### 3.5 Trust Region Policy Optimization (TRPO)

The policy degrade is a difficult problem for all RL methods. TRPO was proposed to boost policy monotonically (Schulman, Levine, et al., 2015). It used KL constraint to ensure small policy update per update. Strictly following monotonic update is impossible while taking into account the calculation expense. Therefore, max KL divergence constraint is replaced by mean KL distance and sampling is used to estimate. TRPO can be written as the following constrained optimization:

$$\min E_{s \sim \rho_{old}, a \sim q}[\frac{\pi_\theta(a|s)}{q(a|s)} Q_{\theta_{old}}(s,a)] \\ s.t \quad E_{s \sim \rho_{old}}[D_{KL}(\pi_\theta(a|s) \| \pi_{\theta_{old}}(a|s))] \leq \delta \quad (7)$$

where $\delta$ is the KL threshold, and $q(a|s)$ is usually set as $\pi_{\theta_{old}}(a|s)$. In fact, uniform can also be used when the action space is discrete.

### 3.6 Proximal Policy Optimization (PPO)

The PPO method was proposed to reduce the complicity of TRPO (Schulman et al., 2017), which is much simpler and easier to be implemented.

Two approaches were proposed, one is moving the constraint to the objectives with adaptive coefficient $\beta$. This constrained optimization can be written as:

$$\max E_{s \sim \rho_{old}}[\frac{\pi_\theta(a|s)}{\pi_{\theta_{old}}(a|s)} A_{\theta_{old}}(s,a) - \\ \beta * KL(\pi_\theta(a|s) \| \pi_{\theta_{old}}(a|s))] \leq \delta . \quad (8)$$

Another approach is clipping the objectives by $\varepsilon$, which can prevent too large policy update. The clipped surrogate objectives can be written as following:

$$\max E_{s \sim \rho_{old}}[\min(r_\theta A_{\theta_{old}}(s,a), \\ clip(r_\theta, 1-\varepsilon, 1+\varepsilon) * A_{\theta_{old}}(s,a)] . \quad (9)$$

Both approaches show good results when applied in single agent continuous problems and were used by OpenAI as their default RL method.

### 3.7 DDPG (Lillicrap et al., 2015)

Instead of outputting stochastic actor policy, DDPG acts deterministically. DDPG updates the policy gradient directly by calculating Q value's semi-gradient, which can be written as:

$$\nabla_\theta J(\mu) = E_{s,a \sim D}[\nabla_\theta \mu(s) * \\ \nabla_a Q(s,a|a=\mu(s))] . \quad (10)$$

DDPG's critic loss is the same as (3).

### 3.8 MADDPG

**MADDPG** uses a centralized critic function taking all the local observation and actions from all agents as input and a decentralized actor function for each agent. The *i*-th agent's policy gradient for the actor part can be written as:

$$\nabla_{\theta_i} J(\mu_i) = E_{x,a \sim D}[\nabla_{\theta_i} \mu_i(o_i) * \\ \nabla_{a_i} Q_i^\mu(x, a_1, ..., a_N) | a_i = \mu_i(o_i))] \quad (11)$$

$\mu_i(a_i | o_i)$ means the action taken by the *i*-th agent under its actor policy when it receives local observations.

Corresponding loss function in the critic part can be written as:

$$L(\theta_i) = E_{x,a,r,x'}[(Q_i^\mu(x, a_1, ..., a_N) - y_i)^2], \quad (12)$$

where

$$y_i = r_i + \gamma Q_i^{\mu'}(x', a_1', ..., a_N')|_{a'_j = \mu'_j(o_j)}. \quad (13)$$

$\mu'$ stands for the target frozen policy network.

## 4. PS deep deterministic policy gradient

### 4.1 Reward sharing and actor-critic sharing

Local observation with imperfect information incurs difficulties for multi-agent cooperation. A common situation is a group of homogeneous agents cooperating to get the job done and meet the requirement for reward sharing. We define reward sharing as following:

$$r_{1t} = r_{2t} = ... = r_{Nt} = r_t. \quad (14)$$

The equation (14) must hold for any time snap,. When reward sharing holds, we use actor and critic sharing (we call PSMADDPG-V0), where all agents share a single actor and critic network. So the policy gradient for the actor can be written as:

$$\nabla_\theta J(\mu) = E_{x,a \sim D}[\nabla_\theta \mu(o_i) * \\ \nabla_{a_i} Q^\mu(x, a_1, ..., a_N) | a_i = \mu(o_i))] \quad (15)$$

The critic loss function can be written as

$$L(\theta_i) = E_{x,a,r,x'}[(Q^\mu(x, a_1, ..., a_N) - y_i)^2], \quad (16)$$

where

$$y_i = r_i + \gamma Q^{\mu'}(x', a_1', ..., a_N')|_{a'_j = \mu'(o_j)}. \quad (17)$$

Strictly speaking, each agent has a different $Q$ function, and uses its own Bellman function to evaluate it. However, when reward-sharing holds, the $Q$ function of each agent is same, which can be easily proven by the definition of $Q$ and we omit it here. In fact, one basic training method is focusing on single agent training. Here we choose a different training process: after sampling from the relay memory, we randomly choose an agent from the agent set, and use this agent to update model parameter based on(15)(16)(17).

Parameter sharing has three explicit advantages over the parameter independent method (the origin MADDPG): 1. the learning speed is nearly N times faster than the latter; 2. the sampled states for training have better diversity, which helps to converge fast and stably; 3. it require nearly N times smaller network memory than the latter.

PSMADDPG-V0 can also be applied when agents meets exchangeability, where reward sharing is not essential. Exchangeability can be defined as following:

$$O_k = O_k', \forall k \neq i \cap k \neq j \\ Set(O_i) = Set(O_j) \quad (18)$$

The equation (18) must hold for any time step. $O_k'$ means local observation for agent *k* after *i* exchange with *j*. The first formula means the other agents' observation stay unchanged after exchange with each other. The second one means two agents have the chance to encounter the same state.

Suppose there is an optimal policy $\mu^*$, each agent behaves as $\mu^*(o)$, and acts based on its local observation $o$. It means each agent should take the same action when they encounter with the same local observation. PSMADDPG-V0 inherently meet this situation. However, the origin MADDPG hardly meets it since different agents' actor networks are trained independently. Agents can still behave with diversity owning to different local observations when applying PSMADDPG-V0.

### 4.2 Non-reward sharing and actor sharing

When reward sharing defined by(14) or (18) does not hold, method using actor sharing MADDPG (PSMADDPG-V1) is proposed.

In such situation, critic functions may be quite different for different agents and it's difficult to learn the macro evaluation function. Therefore different critic functions should be used instead of the shared one.

The critic loss is the same as(12), the actor contains the shared actor network, whose policy gradient can be written as(15).

In fact, there is a variant (which is called V2) between V0 and V1 since it uses shared actor network but partially shared critic with multi-head.

The network architecture is illustrated as Figure 1. All agents share the same actor network, but use a partially shared critic network with multi-head, which is similar to (van Seijen et al., 2017). In critic part, all layers are shared except for the top *n* layer (usually n<3). Head$_i$ is the Q function for agent *i*. This variant fits well in raw pixels input and common part is convolution networks which could learn useful visual features. For raw distilled features, however, it maybe does not perform better than the other two variants.

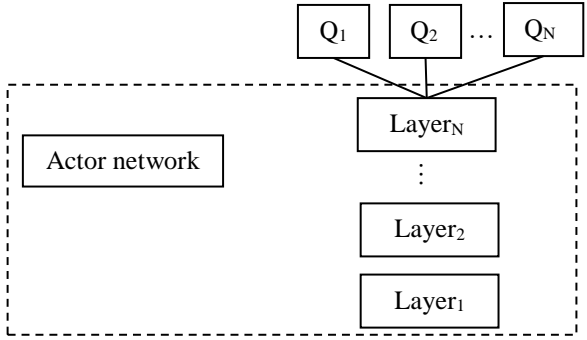

**Figure 1 PSMADDPGV2 architecture. All agents share the same actor network, and partially share the critic network. The dot line rectangle marks the shared part. The N heads correspond with N critic (Q value) function output.**

The actor policy gradient can be written as:

$$\nabla_\theta J(\mu) = E_{x,a \sim D} \frac{1}{N} \sum_{i=1}^{N} [\nabla_\theta \mu(o_i) * \nabla_{a_i} Q^\mu_{head_i}(x, a_1, ..., a_N \mid a_i = \mu(o_i))] \quad (19)$$

The critic loss function can be written as:

$$L(\theta_i) = E_{x,a,r,x'} \frac{1}{N} \sum_{i=1}^{N} [(Q^\mu_{head_i}(x, a_1, ..., a_N) - y_i)^2] , \quad (20)$$

where

$$y_i = r_i + \gamma Q^{\mu'}_{head_i}(x', a'_1, ..., a'_N)\big|_{a'_j = \mu'(o_j)}. \quad (21)$$

$\mu$ is the shared actor network, $\mu'$ is the frozen target network.

## 5. Experiments

Three multi-agent branch[1] benchmarks modified from rllab[2](Duan et al., 2016) are used to test the proposed PSMADDPG variants.

---

[1] https://github.com/rejuvyesh/rllab.git

[2] https://github.com/rll/rllab.git

### 5.1 Setup

For all experiments, we use MADDPG(Lowe et al., 2017), PS-TRPO(Gupta et al., 2017) and PPO(Schulman et al., 2017)as baselines[3].

To make the comparison fairly, we keep all hyper-parameters the same as the origin MADDPG and compare the learning speed and return.

Since it's hard to make such comparison with PSTRPO, we get their published source code and slightly tune hyper parameters.

### 5.2 Water-world

Multi-agent water-world game was extended from the single agent version (Ho et al., 2016). All agents need to cooperate with each other to grasp the food and avoid touching the poisons. Each agent has a local observation with 212 or 213-dimension continuous variables and output 2 continuous force variables along x and y-axis. The observations come from 30 local range sensors with random noise, which estimate the distance with other agents, food target as well as poisons. The difference between 213 and 212 dimension observation lies in whether the agent's id is included. The rewards are composed of various parts, 10.00 for get the food target with cooperation, -1.00 for touching poisons, and extra reward with 0.01 is added when see each other to erase exploration as in (Gupta et al., 2017). The max step for water-world is set to 500.

We test simple situations with 2 agents, 50 food and 50 poisons. The total training steps are set to 2 million. The game terminates when the game reach the max steps within an episode.

Average 100-episode return is illustrated in Figure 2. PSMADDPGV1 and PSTRPO perform better than other methods. MADDPG and PSMADDPGV0 are almost equal. It's not surprising that PPO performs worst when directly applied in multi-agent scenario while it behaves well in single agent scenarios(Schulman et al., 2017). PPO needs some efforts when transferred for multi-agent control. A game playing video could be found here: https://youtu.be/d2SI8DOWB4Q.

We test various methods' sensitivities to hyper parameters, here we only consider learning rate. If learning rate for both networks changes from 1e-4 to 1e-3, MADDPG and V0 degrade drastically but V1 and V2 keep better scores above 100. It's illustrated in Figure 3.

If learning rate for actor network is 1e-4 and learning rate for critic network is 1e-3, all method perform worse. It's illustrated in Figure 4. MADDPG's performance decreases greatly while V1 and V0 still obtain a score above 100. The above experiments shows that MADDPG is more sensitive to hyper-parameters than PSMADDPG.

---

[3] https://github.com/openai/baselines.git. OpenAI release a series of RL algorithms where PPO is also included.

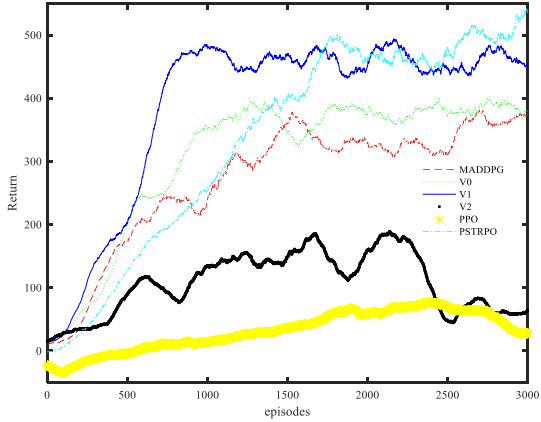

**Figure 2** 2 agents water-world 100 average return for MADDPG and PSMADDPG variants. The learning rate is 0.0001 for both actor and critic networks.

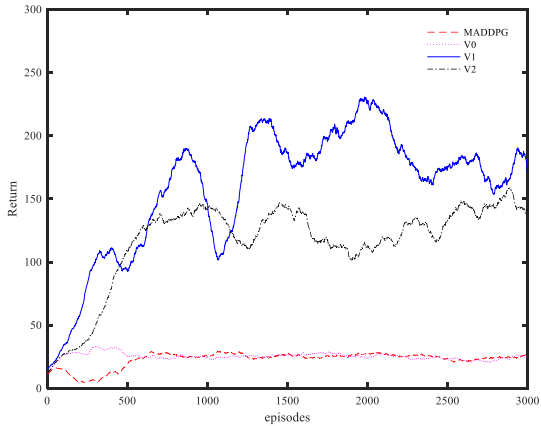

**Figure 3** 2 agents water-world 100 average return for MADDPG and PSMADDPG variants. The learning rate is changed to 0.001 for both actor and critic networks. Note that the PSMADDPG variants can fit better than MADDPG.

### 5.3 Multi-Walker

Multi-walker game is a task in which several agents must cooperate to take a package to a target position. Each agent has a local 32 dimensions observation of front and back partners and offset of the packages with noises and output 4 force torques to control leg joints. The game ends when the package arrives at the target position or any agent falls or package is dropped. The local observations also include the agent id with two alternatives: one hot vector or a float value as $i$/N. As usual, we choose one hot vector. Forward rewards with 1.00 are given by carrying the package forward, and punishments with -5.00 are given when agent falls or package drops. Since multi-walker is more difficult, we run 4 million steps for all methods. Average return for various methods is shown in Figure 5.

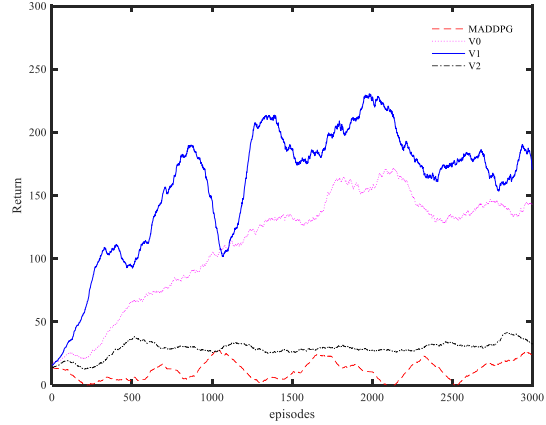

**Figure 4** 2 agents water-world 100 average return for MADDPG and PSMADDPG variants. The learning rate is changed to 0.0001 for actor network and 0.001 for critic network.

PSMADDPGV1 and MADDPG can obtain higher score than others and in fact, they both can do a good job with successful package move. PSTRPO also shows potential high score if given more episodes. Since multi-walker does not meet reward sharing or exchangeability, PSMADDPG V0 does not perform well. We owe V2's bad performance to bad common information distillation. A game playing video could be found here: https://youtu.be/WAtxX-0jrS4.

### 5.4 Multi-Ant

Multi-Ant is extended from mujoco. Each agent is a leg from multi-ant, whose observation contains 18 force sensors, 4 positions, 2 neighboring legs' velocity and 11 world coordinates with noises. The objective is to learn to survive and go as fast as possible. Unlike (Duan et al., 2016), reward for each agent from rllab in the multi-agent version's game can be written as.

$$r_{s,a} = v - C_{contact} - C_{control} + 1.0$$
$$C_{contact} = 5*10^{-4}*\| clip(F,-1,1) \|_2^2 . \quad (22)$$
$$C_{control} = 0.5 \| a \|_2^2$$

The survival reward is 1.0 in rllab while 0.05 is used in(Duan et al., 2016), we guess that the rllab game designer finds it's difficult to survive since the local observations have noises.

Specially, multi-ant game doesn't meet exchangeability. We use six ant legs with decentralized rewards to get a more challenging environment. Our main objective is to verify the effectiveness of PSMADDPG, so we only compare MADDPG and PSMADDPG variants. Another reason is PPO performs badly in the last experiment. The return is shown in Figure 6.

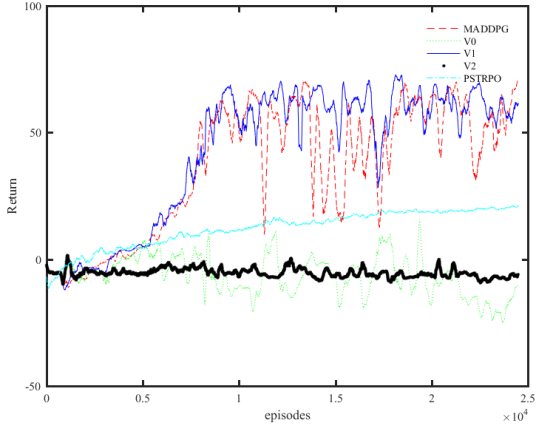

**Figure 5** 2 agents multi-walker 100 average return for MADDPG and PSMADDPG variants. The learning rate is changed to 0.001 for both actor and critic networks.

MADDPG and PSMADDPG variants almost perform equally in multi-ant game and achieve score about 3000. In fact, all agents learns to stand nearly still and move slowly as surviving without risk could obtain higher scores (survival reward is 1.0 every step). We changed the reward to encourage walking forward as far as possible, we decrease the survival reward to 0.05 and change control cost coefficient to 0.005 as (Duan et al., 2016). The reward can be written as:

$$r_{s,a} = v - C_{contact} - C_{control} + 0.05$$
$$C_{contact} = 5*10^{-4} * \| clip(F, -1, 1) \|_2^2 . \quad (23)$$
$$C_{control} = 5*10^{-3} * \| a \|_2^2$$

The experimental results are shown in Figure 7, where V0, V1 and MADDPG score high. As usual, we only plot the total return for all agents for clarity. Although multi-ant doesn't meet reward sharing, V0 can learn to walk fast. PSMADDPG V2 can only learn to stand still, only score about 150. Since the max steps within an episode is 500, each agent can get 0.05*500=25, so the total score is 25*6=150. A game playing video could be found here: https://youtu.be/5Ec_o4Gr9XM.

Parameter sharing variants have two obvious and inborn advantages: low memory requirement and time expense. For each iteration in N-agent cooperation problem, PSMADDPG variants V0 and V2 only update one actor and critic network, V1 update $N$ critic networks and 1 actor network. Meanwhile, MADDPG update $N$ actor and $N$ critic networks. Training time cost can be seen in Table 1.

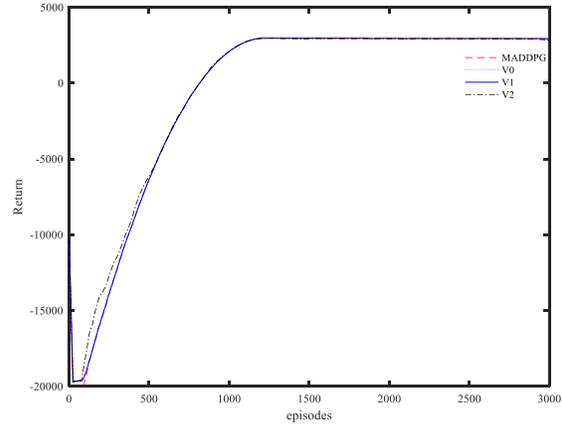

**Figure 6** 6 agents multi-ant 100 average return for MADDPG and PSMADDPG variants

*Table 1* time expense for different methods

|  | WATERWORLD-2 | MULTIWALKER-2 | MULTIANT-6 |
| --- | --- | --- | --- |
| MADDPG | 92364 | 130381 | 221195 |
| V0 | 50743 | 70876 | 57657 |
| V1 | 80693 | 119965 | 130538 |
| V2 | 55336 | 81658 | 107176 |

As Table 1 shows, V0 outperforms other variants, followed by V2, and MADDPG behaves worst as expected.

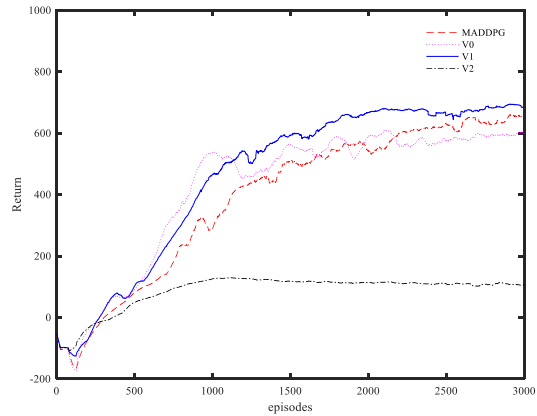

**Figure 7** 6 agents multi-ant 100 average return for MADDPG and PSMADDPG variants after reward redesign.

## 6. Conclusions

We present the parameter sharing deep deterministic policy gradient to handle cooperative multi-agent problem based on MADPPG. Several variants are introduced based on the different sharing strategies. All variants use shared policy network, which makes decisions only based on local observations. The input of critic network contains all agents' observations and actions to make the environment static.

We use benchmarks from rllab to verify the validness of the proposed methods, which show promising better scores and less memory and time expense. More importantly, it can make full use of reward sharing and exchangeability, which is hard to obtain when non-sharing approaches are applied.

The previous work focuses on cooperative homogeneous agents. However, the presented methods in this paper can be easily extended to competitive situations where cooperative homogeneous agents share single actor network, and heterogeneous agents use different actor networks. It makes this method more scalable when applied for situations with many agents, where most of existing approaches including MADDPG fail.

Our further work will focus on extending the proposed method to more complex situations with a large amount of agents such as StarCraft.

# Appendix

To make consistent representations, we use the same symbols as (Lowe et al., 2017), and mark the differences in bold when necessary.

**Algorithm 1** PSMADDPGV0

**Given sharing** actor $\theta_a$ and critic $\theta_c$, target actor $\theta'_a$ and critic $\theta'_c$ network for all agents.
Initialize actor and critic network parameters.
**for** episode 1 to M do
   Initialized a random process $\mathcal{N}$ for action exploration
   Receive initial state x
   **for** t=1 to max-episode-length **do**
     for each agent $i$, select action $a_i = u_{\theta_a}(o_i) + N_i$
     Execute actions $a = (a_1,...,a_N)$ and receive reward r and next new state $x'$
     Store $(x,a,r,x')$ into replay memory $D$
     $x \leftarrow x'$
   **Randomly choose an agent $i$**
     Sample a random mini-batch of $S$ samples $(x^j, a^j, r^j, x'^j)$ from $D$
     Set $y_i^j = r_i^j + \gamma Q^{\mu_{\theta_c}}(x'^j, a'_1,...,a'_N)|_{a'_j = \mu_{\theta_a}(o_j)}$
     Update critic by minimizing the loss by $L(\theta_c) = \frac{1}{S}\sum_j [(Q^{u_{\theta_c}}(x,a_1,...,a_N) - y_i)^2]$
     Update actor by using sampling policy gradient
$$\nabla_{\theta_a} J(\mu) \approx \frac{1}{S}\sum_j [\nabla_{\theta_a}\mu(o_i)\nabla_{a_i}Q^{\mu}(x,a_1,...,a_N | a_i = \mu_{\theta_a}(o_i))]$$

   **Update target parameter.**
$$\theta'_c = \tau * \theta_c + (1-\tau) * \theta'_c$$
$$\theta'_a = \tau * \theta_a + (1-\tau) * \theta'_a$$

**Algorithm 2** PSMADDPGV1

**Given sharing** actor $\theta_a$ and target actor $\theta'_a$, non-sharing critic $\theta_{ci}$ and $\theta'_{ci}$ network for all agents,
Initialize actor and critic network parameters.
**for** episode 1 to M do
   Initialized a random process $\mathcal{N}$ for action exploration
   Receive initial state x
   **for** $t$=1 to max-episode-length **do**
     for each agent $i$, select action $a_i = u_{\theta_a}(o_i) + N_i$
     Execute actions $a = (a_1,...,a_N)$ and receive reward r and next new state $x'$
     Store $(x,a,r,x')$ into replay memory $D$
     $x \leftarrow x'$
   **For each agent $i$ do**
     Sample a random mini-batch of $S$ samples $(x^j, a^j, r^j, x'^j)$ from $D$

Set $y_i^j = r_i^j + \gamma Q^{\mu_{\theta_{ci}}}(x'^j, a_1', ..., a_N')|_{a'_j = \mu_{\theta_a}(o_j)}$

Update critic by minimizing the loss by $L(\theta_c) = \frac{1}{S}\sum_j [(Q^{u_{\theta_{ci}}}(x, a_1, ..., a_N) - y_i)^2]$

Update actor by using sampling policy gradient

$$\nabla_{\theta_a} J(\mu) \approx \frac{1}{S}\sum_j [\nabla_{\theta_a} \mu(o_i) \nabla_{a_i} Q^\mu(x, a_1, ..., a_N | a_i = \mu_{\theta_a}(o_i))]$$

**Update actor target parameter**

$$\theta_a' = \tau * \theta_a + (1-\tau) * \theta_a'$$

**Update critic target parameter for each agent**

$$\theta_{ci}' = \tau * \theta_{ci} + (1-\tau) * \theta_{ci}'$$

---

**Algorithm 3** PSMADDPGV2

---

**Given sharing** actor $\theta_a$ and critic $\theta_c$, target actor $\theta_a'$ and critic $\theta_c'$ network for $N$ agents.
Initialize actor and critic network parameters.
**for** episode 1 to M do
   Initialized a random process $\mathcal{N}$ for action exploration
   Receive initial state x
   **for** t=1 to max-episode-length **do**
     for each agent $i$, select action $a_i = u_{\theta_a}(o_i) + \mathcal{N}_i$
     Execute actions $a = (a_1, ..., a_N)$ and receive reward r and next new state $x'$
     Store $(x, a, r, x')$ into replay memory $D$
     $x \leftarrow x'$
     Sample a random mini-batch of $S$ samples $(x^j, a^j, r^j, x'^j)$ from $D$
     Set $y_i^j = r_i^j + \gamma Q^{\mu_{\theta_c}}(x'^j, a_1', ..., a_N')|_{a'_j = \mu_{\theta_a}(o_j)}$
     Update critic by minimizing the loss by $L(\theta_c) = \frac{1}{NS}\sum_{i=1}^{N}\sum_{j=1}^{S}[(Q^{u_{\theta_c}}(x, a_1, ..., a_N) - y_i)^2]$
     Update actor by using sampling policy gradient

$$\nabla_{\theta_a} J(\mu) \approx \frac{1}{NS}\sum_{i=1}^{N}\sum_{j=1}^{S}[\nabla_{\theta_a}\mu(o_i)\nabla_{a_i}Q^\mu(x, a_1, ..., a_N | a_i = \mu_{\theta_a}(o_i))]$$

**Update target parameter.**

$$\theta_c' = \tau * \theta_c + (1-\tau) * \theta_c'$$
$$\theta_a' = \tau * \theta_a + (1-\tau) * \theta_a'$$

---

### Experiments Details

For MADDPG, PSMADDPGV0, PSMADDPGV1 experiments, the actor network contains three fully-connected (FC) layers (500, 128, action dimensions) and activate function is used for the first two layers, tanh is used for the last layer. The critic network contains four layers FC (500, 300, 128), relu activate function is used for the first three layers, and linear is used without activate function for the last layer. In addition, we set a same seed for each methods within same episode number, but different for different episodes.

In PSMADDPGV2 experiments, the actor network has the same setting as above. And for multi-head critic network, the shared part is composed of two FC layers (500,300), and each head is compose of one hidden FC layer(128) followed by one output neutron. For all experiments except for learning rate comparison, we use Adam optimizer with learning rate 0.0001. For all experiments, we use discount $\gamma=0.99$.

Exploration epsilon decreases linearly from 1 to 0.02 within the first 600000 steps and stays unchanged until to end in all experiments.

For PSTRPO, we use the same MLP setup as (Gupta et al., 2017): three FC layers (100,50,25), KL constraint step limit $\delta$ is set to 0.01 and batch size limit is 24000.

For PPO, we use the two hidden layers (256,256), 2048 time step, 0.0003 step size and 64 optimal batch size. Moreover, GAE(0.95)(Schulman, Moritz, et al., 2015) is use to estimate advantage.

We made all experiments on NVIDIA Tesla P100.